\documentclass[runningheads]{llncs}

\usepackage[T1]{fontenc}
\usepackage{graphicx}
\usepackage{comment}
\usepackage{amsmath,amssymb}
\usepackage{color}
\usepackage{url}
\usepackage{hyperref}
\usepackage{algorithm}
\usepackage{algorithmic}
\usepackage{epstopdf}

\newif\ifreview
\reviewfalse

\ifreview
	\usepackage{lineno}

	\linenumbers
\fi

\begin{document}

%%%%%%%%%%%%%%%%%%%%% Add submission id, track, and title. %%%%%%%%%%%%%%%%%%%%%

\def\SubNumber{098}

\def\GCPRTrack{Pattern recognition in the life and natural sciences}

\title{Learning Channel Importance for High Content Imaging with Interpretable Deep Input Channel Mixing}
% You can use \thanks for acknowledgment as in: 
%\title{Title\thanks{XXX}}
%Do not add any acknowledgment to the draft 
% version that is used for the review process.  

\ifreview
	% ANONYMOUS SUBMISSION FOR REVIEW
	% DO NOT MODIFY these for the draft version that is used for the review process.
	\titlerunning{GCPR 2023 Submission \SubNumber{}. CONFIDENTIAL REVIEW COPY.}
	\authorrunning{GCPR 2023 Submission \SubNumber{}. CONFIDENTIAL REVIEW COPY.}
	\author{GCPR 2023 - \GCPRTrack{}}
	\institute{Paper ID \SubNumber}
\else

	\author{Daniel Siegismund \and
		Mario Wieser \and
		Stephan Heyse \and
		Stephan Steigele}
	
	\authorrunning{D. Siegismund, M. Wieser et al.}
	\titlerunning{DCMIX}
	
	\institute{Genedata AG, Basel, Switzerland}
\fi

\maketitle              % typeset the header of the contribution

\begin{abstract}
	Uncovering novel drug candidates for treating complex diseases remain one of the most challenging tasks in early discovery research. To tackle this challenge, biopharma research established a standardized high content imaging protocol that tags different cellular compartments per image channel. In order to judge the experimental outcome, the scientist requires knowledge about the channel importance with respect to a certain phenotype for decoding the underlying biology. In contrast to traditional image analysis approaches, such experiments are nowadays preferably analyzed by deep learning based approaches which, however, lack crucial information about the channel importance. To overcome this limitation, we present a novel approach which utilizes multi-spectral information of high content images to interpret a certain aspect of cellular biology. To this end, we base our method on image blending concepts with alpha compositing for an arbitrary number of channels. More specifically, we introduce DCMIX, a lightweight, scaleable and end-to-end trainable mixing layer which enables interpretable predictions in high content imaging while retaining the benefits of deep learning based methods. We employ an extensive set of experiments on both MNIST and RXRX1 datasets, demonstrating that DCMIX learns the biologically relevant channel importance without scarifying prediction performance.
\end{abstract}
\section{Introduction}
\label{introduction}
High-Content Imaging (HCI) has developed to one of the main driving factors in biopharma early discovery research to reveal novel drug candidates for sophisticated treatment strategies such as cancer immunotherapies \cite{cancerim}. HCI is based on a standardized experimental protocol that allow for the systematic acquisition of multi-spectral images, e.g., in form of a cell painting assay protocol that requires a high number of channels with the benefit of a highly generalizable assay \cite{bray_cell_2016}. Here, high-content images are recorded by automated instruments on microtiter plates which allow for large-scale drug candidate testing and an automatic analysis procedure to assess the mechanics of a drug candidate for a certain disease. When running such HCI experiments, scientists prepare typically a set of 4 to 15 channels \cite{levenson_multispectral_2006,nalepa_recent_2021} with a specific fluorophore that tags a certain cellular protein or compartment. Subsequently, the scientist aims to analyze the experimental outcome with respect to the importance of the fluorescence channels to validate the findings or refine the experiment and, therefore, requires a fast and easy-to-use analysis workflow. This is particularly important as the specific functional or mechanistic knowledge is encoded via the specific staining per image channel \cite{bray_cell_2016} and hence required for decoding the underlying biology.

However, to analyze such complex multi-channel cell-painting assays, the scientist requires the ability of sophisticated image analysis to distill the information from the multi-spectral information. In biopharma research, the traditional analysis \cite{carpenter_cellprofiler_2006} is gradually replaced by deep learning based approaches \cite{steigele_deep_2020,godinez_multi-scale_2017,9911974,pmlr-v172-siegismund22a}. Despite the superior performance of such models in comparison to conventional segmentation based analysis \cite{carpenter_cellprofiler_2006}, the scientist lacks informative insights in terms of understanding about which fluorescence channel influenced the decision \cite{castelvecchi_can_2016}.

In the past, various approaches have been proposed to extract the most relevant information from high-dimensional datasets. The most basic approach to determine the most relevant channels is a preprocessing step by applying an unsupervised dimensionality reduction method such as Principal Component Analysis (PCA) \cite{Jolliffe}. However, employing such a preprocessing step does not guarantee for phenotype-specific channels as the method only optimizes for the directions with the highest variance and not necessarily for the highest phenotypic information. More recently, attention-based approaches have been introduced for image channel selection \cite{cai_bs-nets_2020,he_dual_2022,li_attention_2022,nikzad_attention-based_2022} which suffer from high computational costs and poor scalability. In addition, there are model-agnostic approaches such as Shapely values \cite{rozemberczki_shapley_2022,jullum_groupshapley_2021} which, however, can suffer from sampling variability \cite{lundberg2020local2global} and be time consuming in terms of highly complex models \cite{Carrillo}.

To overcome the aforementioned limitations, we present a simple yet effective method to estimate channel importance for HCI images. More specifically, we introduce a lightweight, easy to use mixing layer that is composed of a generalized image blending mechanism with alpha compositing \cite{zhang_deep_2020,kokalj_why_2019} which converts a $d$-dimensional channel image into a 2D image retaining all phenotype relevant information. This allows not only to incorporate an arbitrary number of channels in a highly scalable fashion but also leads to a reduced network size with faster inference times while being able to facilitate the use of transfer learning of pretrained networks.
To summarize, we make the following contributions:
\begin{itemize}
	\item We extend the imaging blending concepts of  \cite{zhang_deep_2020,kokalj_why_2019} and apply these to images with an arbitrary number of channels.
	\item We encapsulate the generalized image blending into a lightweight, scalable and end-to-end trainable mixing layer, called DCMIX, to estimate channel importance for multi-spectral HCI data.
	\item Experiments on MNIST as well as on the challenging multi-channel real-world imaging data set RXRX1 \cite{sypetkowski_rxrx1_2023} with 31 different cell phenotype classes demonstrate that the proposed method learns the correct channel importance without sacrificing its model performance.
\end{itemize}

\section{Related Work}
\label{related_work} 
In this section, we review related work on interpretable and explainable machine learning \cite{widm.1493}. Broadly spoken, we can distinguish between interpretable models that are interpretable by design and explainable models that try to explain existing models post-hoc \cite{widm.1493}.

\subsubsection{Interpretable Machine Learning Methods} can be separated into the following model classes:  score-based \cite{Ustun_2015}, rule-based \cite{Cohen95}, sparse \cite{tibshirani96regression} and neural networks \cite{feng2019sparseinput}, among others \cite{widm.1493}. In this review, we focus more closely on sparsity inducing and attention-based interpretable methods. Sparsity-based approaches introduce a sparsity constraint on the model coefficients to determine the feature importance. One of the most basic approaches is the least absolute shrinkage and selection operator (LASSO) introduced by \cite{tibshirani96regression} which is employing the $L_1$-norm to ensure feature sparsity. This approach has subsequently been extended to various lines of research, including dealing with grouped features \cite{Ming,park_bayesian_2008}, estimating network graphs \cite{friedman_sparse_2008,prabhakaran12} or learning sparse representations in neural networks \cite{louizos2018learning,wieczorek2018learning}. Most closely related to our work is LassoNet \cite{JMLR:v22:20-848} which employs a group lasso constraint based on the feature channels that are obtained from a pretrained feature extraction network. In contrast, our approach is end-to-end trainable and hence does not require a two step approach of feature extraction and importance estimation. More recently, attention-based approaches \cite{NIPS2017_3f5ee243} have emerged in the context of interpretable machine learning. \cite{choi2017retain} introduced an attention-based model for the analysis of electronic health records and \cite{schwab2019ame} learns important features with an attentive mixture of experts approach. Moreover, attention is used in the context of hyper spherical band/channel selection \cite{cai_bs-nets_2020,he_dual_2022,li_attention_2022,nikzad_attention-based_2022}. In contrast, our approach works on image blending and alpha compositing and hence reducing high computational costs.

\subsubsection{Explainable Machine Learning Methods}
\label{rel:shapley}
denote approaches that aim to explain decisions of an already trained machine learning model post-hoc by learning a surrogate model \cite{widm.1493}. In summary, we distinguish between attribution methods that try to quantify the attribution of a feature to the prediction \cite{pmlr-v70-sundararajan17a}, concept-based explanations trying to explain predictions with high-level concepts \cite{pmlr-v80-kim18d}, symbolic metamodels employing symbolic regression as a surrogate \cite{NEURIPS2019_567b8f5f} and counterfactual explanations \cite{Wachter2017CounterfactualEW}. In the context of our work, we focus on attribution models. \cite{ribeiro2016why} learns a surrogate classifier to explain an arbitrary black-box model based on submodular optimization. \cite{pmlr-v70-shrikumar17a} introduced DeepLIFT to decompose the input contributions on the model prediction. In addition, Shapley values gained a wide adoption in the machine learning domain mainly for feature selection and model explain ability \cite{rozemberczki_shapley_2022,jullum_groupshapley_2021}. As a result, Lundberg \& Lee \cite{NIPS2017_8a20a862} introduced Shapely additive explanations (SHAP) to explain model predictions based on Shapely regression values. Finally, Shapley values have been used in the context of HCI channel importance estimation \cite{sypetkowski_rxrx1_2023}. More specifically, the authors adopt Shapely values to explain the channel importance of HCI images from a pretrained black-box model. Opposed to our approach, this method requires the training of two separate models and hence does not allow for end-to-end training.

\begin{figure}[!th]
	\centering
	\includegraphics[width=0.8\textwidth]{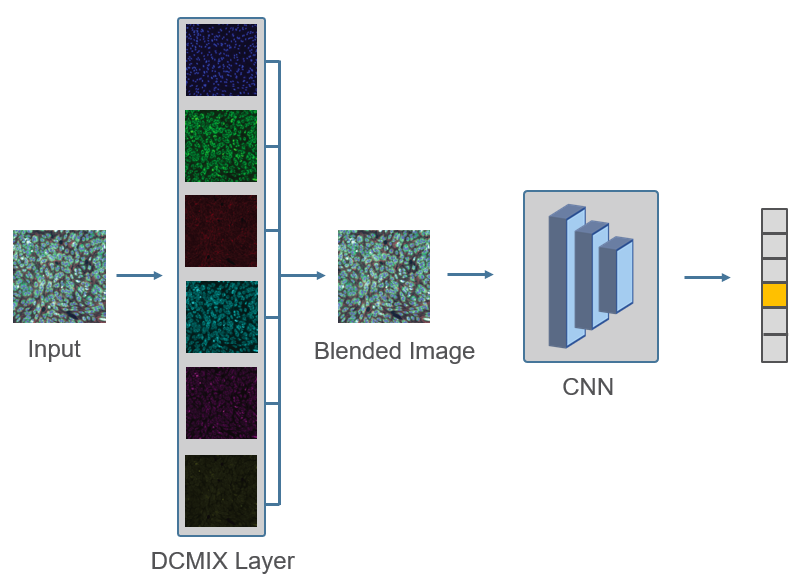}
	\caption{Blue arrows denote steps and gray boxes actions in our workflow, respectively. In the first step
		(1.), we take a multi-channel cellular image and split it into single channels. Subsequently, we mix the channel within our DCMIX layer to obtain the most important part of each channel. In the second step (2.), we take the blended image into our classification network.}
	\label{overview}
\end{figure}

\section{Model}
As illustrated in Figure \ref{overview}, we utilize a two step approach for estimating channel importance in multi-spectral bioimage classification settings by introducing a lightweight,
easy to use and end-to-end trainable mixing layer. To do so, we propose a blending layer which combines the most important parts of the distinct
channels into a new 2D image. After, we perform a classification based on the blended image.

\subsection{Conception of the Image Blending Layer}
We start with an input image $I \in \mathbb{R}^{h \times w \times c}$ where $h$ denotes the height, $w$ the width and $c$ the number of channels in the multi-spectral image. Subsequently, the image $I$ is split into its distinct channels and processed in the DCMIX layer. The DCMIX layer is inspired by simple image blending and alpha compositing \cite{zhang_deep_2020,kokalj_why_2019}.
More specifically, the idea behind Alpha blending is to combine two images as follows:

\begin{equation}
	C = \alpha_1 \cdot A_1 + (1-\alpha_1) \cdot A_2,
	\label{eq:alpha}
\end{equation}

\noindent where $`A_1 \in` \mathbb{R}^{h \times w \times c}$ and $A_2  \in \mathbb{R}^{h \times w \times c}$ are the corresponding image matrices to blend and $C  \in \mathbb{R}^{h \times w \times c}$ the blended image matrix. The trainable parameter $\alpha_1$ determines the transparency of each channel.

In this work, we take advantage of the ideas proposed in \cite{zhang_deep_2020,kokalj_why_2019} and generalize the idea by employing the trainable
alpha values as weights for each channel that has to be blended with:

\begin{equation}
	C = \sum^n_i \alpha_i \cdot A_i
	\text{ where:   }
	\alpha_i \geq 0,
	\label{eq:DCMIX}
\end{equation}
where $\alpha_i$ is multiplied with each channel $A_i$. The parameter $n$ defines the number of channels and $C$ is the blended image which will be subsequently used for the further analysis.

\subsection{Classifying Genetic Perturbations based on DCMIX-blended Images }\label{sec:networks}
Our goal is to learn a classification model $F_\theta(y \mid C)$ of the blended image $C$ for distinct classes of genetic perturbations $y^c$ where $c$ is the number of genetic perturbations to be predicted. In this work, our model $F$ is a Deep Convolutional Neural Network which extracts a cascade of feature maps $M^l$ where $l$ denotes the current layer. The last feature map is used as an input to the multi-class classification head that predicts the genetic perturbation vector $y^c$ using a softmax output.

\subsection{End-to-End Training Algorithm}
The model training is described in Algorithm \ref{algo1}. As an input, we use the multispectral images $X$ and the genetic perturbation labels $y$. Subsequently, we draw minibatches from the training data $X,y$ (line 1). For each of the minibatches, we obtain the blended images $c_i$ as well as the corresponding mixing factors $\alpha_i$. The blended images $c_i$ are fed in the neural network $F_\theta$ (line 3) and the corresponding predictions $\hat{y_i}$ are used to calculate the loss in line 5. Finally, we update the parameters $\theta$ and $\alpha$ based on the loss by using gradient descent (line 7).

\begin{algorithm}
	\caption{DCMIX training algorithm}
	\label{algo1}
	\begin{flushleft}
		\textbf{INPUT:} $X$ images, $y$ labels\\
		\textbf{OUTPUT:} The prediction $\hat{y}$, mixing factors $\alpha$
	\end{flushleft}
	\begin{algorithmic} [1]
		\FOR{minibatch $x_i, y_i$ from $X, y$} 
		\STATE $c_i, \alpha_i \leftarrow \text{DCMIX}(x_i)$
		\STATE $\hat{y_i} \leftarrow F_\theta(c_i)$
		\STATE 
		\STATE $\text{loss} \leftarrow \text{crossentropy}(\hat{y_i}, y_i)$
		\STATE 
		\STATE update $\theta, \alpha$ using gradient descent
		\ENDFOR
	\end{algorithmic}
\end{algorithm}

\section{Experiments}

A description of setups and additional hyperparameters can be found in the supplementary materials.

\subsection{MNIST}

\subsubsection{Dataset.}
To demonstrate the efficacy of DCMIX for estimating channel importance, we generate an artificial dataset based on MNIST \cite{deng2012mnist}. MNIST consists of 70000 samples with images $x \in  \mathbb{R}^{28x28x1}$ and labels $y$ that represent numbers from 0 to 9. For our dataset, we randomly select a subset of 10000 samples from MNIST. In order to assess the channel importance, we extend the MNIST images with two additional noise channels. Therefore, we draw two noise matrices with shape $28x28$ from a uniform distribution defined on $[0,255]$. Subsequently, we add the previously generated noise channels to the input image such that we obtain a three channel input image $x \in  \mathbb{R}^{28x28x3}$ where the first denotes the most important channel. For training, we split the data into a 70 percent training and a 30 percent hold-out set. The training set is further split into a 80 percent training and 20 percent validation set, respectively.

\subsubsection{Models.} In order to demonstrate the effectiveness of our approach, we benchmark DCMIX against a plain LCNet050, LassoNet \cite{JMLR:v22:20-848} as well as on an attention-based \cite{luong_effective_2015,lin_structured_2017} LCNet050.

\subsubsection{Quantitative Evaluation. Channel Importance}
In this experiment, we evaluate the channel importance on the validation set, and the results are reported in Table \ref{tab:ch:mnist}. As we can observe in the channel importance ranking, DCMIX can effectively learn the most important channel one and is in line with the more complex LassoNet and attention-based LCNet050. At the same time, DCMIX requires only a fraction of GFLOPS and model parameters. More specifically, DCMIX requires solely 5.9271 GFLOPS compared to 17.809 GFLOPS for the Attention-LCNet050. In addition, DCMIX need three times less parameters (0.2789 million) in contrast to Attention-LCNet050 (0.9281 million) and requires only the same amount of GFLOPS and parameters as the plain LCNet050.

\begin{table}
	\caption{Results of the MNIST channel importance and model size. Channel importance ranking denotes the rank of the weights depicted in the second column. The model size is evaluated on GFLOPS and the number of model parameters where lower is better.}
	\label{tab:ch:mnist}
	\centering
	\begin{tabular}{|p{2.5cm}|p{1.5cm}|p{4cm}|p{1.6cm}|p{2.0cm}|}
		
		\hline
		Method & Channel importance ranking & Channel weights & GFLOPS & \# Parameters (million)\\
		\hline
		LCNet050 & - & - & 5.9269 & 0.2789\\
		LassoNet \cite{JMLR:v22:20-848} & 1,3,2 & 120259, 51003, 52318 & - & -\\
		Attention\cite{luong_effective_2015,lin_structured_2017}-LCNet050 & 1,3,2 & 1,$3.24 \times 10^{-11}$, $2.33 \times 10^{-6}$ & 17.809 & 0.9281\\\hline \hline
		\textbf{DCMIX-LCNet050} &  1,3,2 & 0.82,0.21,0.22 & 5.9271  & 0.2789\\
		\hline
	\end{tabular}
\end{table}

\subsubsection{Quantitative Evaluation. Model Performance}
Despite the fact, that the aim of this method is not to improve the model performance but rather learn the most important channel to gain biological insights for a drug discovery experiment, we want to ensure that DCMIX archives competitive performance to state-of-the art approaches. To do so, we compared DCMIX to a plain LCNet050, LassoNet and Attention-LCNet050 in Table \ref{tab:pred:mnist}. Here, we observe that DCMIX obtains competitive results compared to both LCNet050 and Attention-LCNet050 and outperforms LassoNet on accuracy, precision, recall and f1-score measures.

\begin{table}
	\caption{Results of model performance for the MNIST dataset on the hold-out dataset. We assess the model performance on four different metrics: accuracy, precision, recall and f1-score where higher is better. Values in brackets denote the standard deviation.}
	\label{tab:pred:mnist}
	\centering
	\begin{tabular}{|p{3.5cm}|p{2cm}|p{2cm}|p{2cm}|p{2cm}|}
		
		\hline
		Method & Accuracy & Precision & Recall & F1-Score\\
		\hline
		LCNet050 & 0.992 (0.0008) & 0.991 (0.002) & 0.991 (0.002) & 0.991 (0.002)\\
		LassoNet \cite{JMLR:v22:20-848} & 0.963 (0.012) & 0.888 (0.002) & 0.888 (0.002) & 0.887 (0.002)\\
		Attention\cite{luong_effective_2015,lin_structured_2017}-LCNet050 & 0.992 (0.002) & 0.991 (0.001) & 0.991 (0.001) & 0.991 (0.001)\\\hline \hline
		\textbf{DCMIX-LCNet050} &  0.991 (0.002) & 0.990 (0.002) & 0.990 (0.002)  & 0.990 (0.002)\\
		\hline
	\end{tabular}
\end{table}

\subsection{RXRX1}
\subsubsection{Dataset.}
\label{data}
For our real world experiment, we employ the RXRX1 dataset\cite{sypetkowski_rxrx1_2023} which consists of 125510 512x512 px fluorescence microscopy images (6 channels) of four different
human cell lines that are perturbed with 1138 genetic perturbations (including 30 different positive control perturbations). In this study, we used as the training data 30 positive control siRNAs plus the non-active control which lead to 31 classes in total. All images were normalized using the 1 and 99 percent percentile and after, we extract image patches with a size of 192x192 px and an offset of 96 px. This step leads to 32776 image patches. For training, we split the data into a 70 percent training and a 30 percent hold-out set. The training set is further split into a 80 percent training and 20 percent validation set, respectively.

\subsubsection{Models} For the real-world RXRX1 experiment, we compare DCMIX to LassoNet \cite{JMLR:v22:20-848} and the attention-based \cite{luong_effective_2015,lin_structured_2017} LCNet050.

\begin{table}
	\caption{RXRX1 channel importance evaluation for the HepG2 cell line. The importance ranking illustrates the most important channels form left to right based on the weights depicted in the second column. In addition, model statistics are measured in GFLOPS and the number of model parameters (lower is better).}
	\label{tab1}
	\centering
	\begin{tabular}{|p{3.2cm}|p{1.5cm}|p{4cm}|p{1.5cm}|p{1.5cm}|}
		
		\hline
		Method &  Importance ranking & Channel weights (in Channel order) & GFLOPS & \# parameters (millions)\\
		\hline		
		ViT-B16-Imagenet21k \cite{dosovitskiy_image_2020} + LassoNet \cite{JMLR:v22:20-848} & 6,4,1,5,2,3 & 73084, 52526, 31138, 87881, 55612, 107733 & - & -\\
		
		Attention-LCNet050 & 4,2,5,1,3,6 & 0.15, 0.17, 0.008, 0.48, 0.16, 0.007 & 35.61 & 1.75\\\hline \hline
		
		\textbf{DCMIX-LCNet050} & 4,2,3,5,1,6 & 0.30, 0.69, 0.38, 1.06, 0.36, 0.21 & 5.95 & 0.27\\
		\hline
	\end{tabular}
\end{table}

\subsubsection{Quantitative Evaluation. Channel Importance}
Here, we describe the evaluation results on channel importance for the RXRX1 dataset which
is illustrated in Table \ref{tab1}. To do so, we compare the results to the ground truth
introduced in \cite{sypetkowski_rxrx1_2023}. The experiment was manually designed by a
scientist in the laboratory such that both channels four and two hold the most important
biological information and channel 6 contains no important information for the phenotype.
Keeping this information in mind, we assess the channel importance of DCMIX, LassoNet and Attention-LCNet050.
Here, we can confirm that DCMIX learns the two most important channels four and two and the least important channel 6.
These findings are also supported by Attention-LCNet050 which learned equivalent importance values.
In contrast, LassoNet fails to uncover the correct channel importance by selecting the least important channel as
the most important one. Despite finding the same important channels, DCMIX possess a 6-8 times higher speed
and requires 6 times less parameters compared to the attention based networks and can be used in an end-to-end
fashion which is not feasible for LassoNet.

\subsubsection{Quantitative Evaluation. Model Performance} In this experiment, we evaluate the model performance of
DCMIX to LassoNet and Attention-LCNet050 and illustrate the results in Table \ref{tab:pred:rxrx}.
Here, we observe that DCMIX outperforms both LassoNet and Attention-LCNet050 in terms of accuracy by five and seven percent,
respectively. Furthermore, these finding are confirmed by precision, recall and f1-scores where DCMIX outperforms
both competitors by approximately five and seven percent.

\begin{table}
	\caption{Results of model performance for the RXRX1 dataset on the hold-out dataset. We asses the model performance on four different metrics: accuracy, precision, recall and f1-score where higher is better. Values in brackets denote the standard deviation.}
	\label{tab:pred:rxrx}
	\centering
	\begin{tabular}{|p{3.5cm}|p{2cm}|p{2cm}|p{2cm}|p{2cm}|}
		
		\hline
		Method & Accuracy & Precision & Recall & F1-Score\\
		\hline
		ViT-B16-Imagenet21k \cite{dosovitskiy_image_2020} + LassoNet \cite{JMLR:v22:20-848} & 0.695 (0.004) & 0.705 (0.005) & 0.705 (0.004) & 0.704 (0.005)\\
		Attention\cite{luong_effective_2015,lin_structured_2017}-LCNet050 & 0.744 (0.019) & 0.753 (0.014) & 0.747 (0.014) & 0.747 (0.013)\\\hline \hline
		\textbf{DCMIX-LCNet050} &  0.765 (0.004) & 0.77 (0.037) & 0.77 (0.042)  & 0.764 (0.043)\\
		\hline
	\end{tabular}
\end{table}

\begin{figure}[t]
	\centering
	\includegraphics[width=0.9\textwidth]{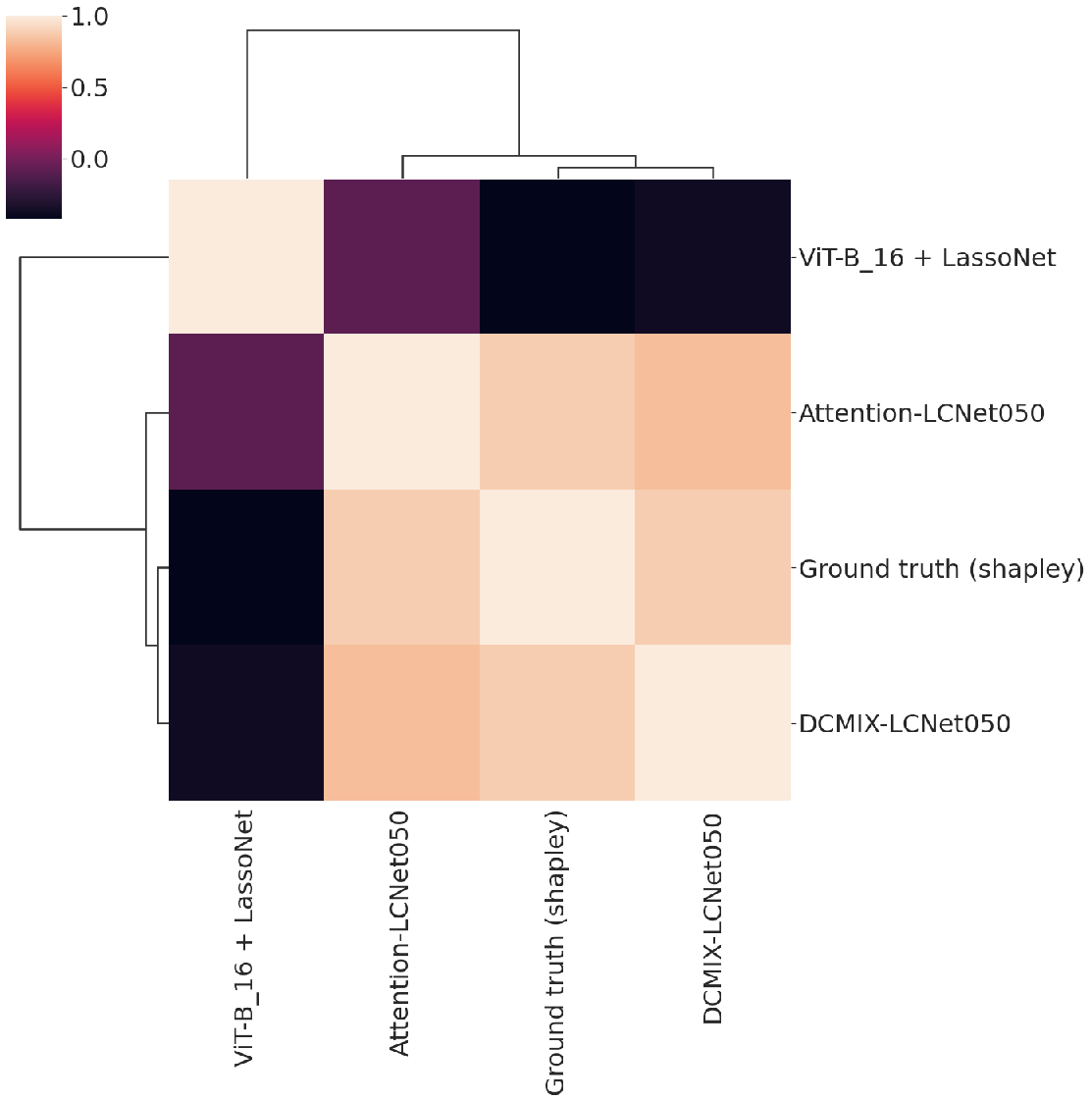}
	\caption{Visualization of Spearman's rank correlation coefficient of the channel importance estimates for all different methods from Table \ref{tab1}.
		A value of -1 indicates maximal ranking difference between the channel importance estimates, 1 indicates no difference.
		Matrix has been sorted using average linkage hierarchical
		clustering with euclidean distance.}
	\label{fig2}
\end{figure}

\section{Discussion}

\subsubsection{DCMIX demonstrates state-of-the-art channel importance scores in fluorescence cellular imaging} DCMIX employs
image blending to estimate the importance of each image channel.
In Figure \ref{fig2}, we provide an overview of the Spearman rank correlation of the channel importance
estimates for all tested methods.
The results are comparable for all methods (except of LassoNet) with a Spearman $\rho$ always larger than 0.83.
Especially the correlations of DCMIX and Attention-LCNet050 to the ground truth shapley values from \cite{sypetkowski_rxrx1_2023} are
evident with a Spearman $\rho$ of 0.89.
Both methods estimate the channel 2 and 4 as most important and channel 6 as least important which was the intentionally experimental design
and furthermore shown via shapley values \cite{sypetkowski_rxrx1_2023}.
The authors explained their finding with a very large spectral overlap of the fluorescence signal from channel 2 and 4
to any other channel rendering them more important \cite{sypetkowski_rxrx1_2023}.\\
In contrast, LassoNet does not show any overlap with the rankings selected by all other methods (Figure \ref{fig2})
with a maximal Spearman $\rho$ value of -0.08.

\subsubsection{DCMIX achieves state-of-the-art classification performance with lower model complexity} Across all classification metrics DCMIX archives competitive results on MNIST and state-of-the-art performances on real-world RXRX1 compared to its competitors. Intuitively, we attribute the competitive results on MINST to the problem simplicity which is further supported by the high classification scores of 99\% (Table \ref{tab:pred:mnist}). Concurrently, DCMIX requires merely a fraction of model parameters in all experiments (Tables \ref{tab:ch:mnist},\ref{tab1}) compared to the baselines. 

\subsubsection{Practical runtimes for DCMIX are 6-8 times faster than Attention-based approaches} While DCMIX requires only 5.9271 GFLOPS and 5.95 GFLOPS on RXRX1, achieving the same computational performance as plain LCNet050, Attention-LCNet050 needs 17.809 GFLOPS on MNIST and 35.614 GFLOPS on RXRX1, respectively (Table \ref{tab:ch:mnist} and Table \ref{tab1}). Moreover, even post-hoc approaches such Shapely values that are trained on a black-box model require often more significant computation time. For example, the training time required for the Shapley value explanation are in the range of several minutes for the smaller CIFAR-10 dataset \cite{jethani_fastshap_2022}. This demonstrates that the speed of DCMIX outperforms not only interpretable competitors but also explainable post-hoc approaches on a large scale.

\subsubsection{DCMIX is applicable in real-world settings beyond biomedical imaging} From an application standpoint we see an advantage of DCMIX over the other tested methods, as the high scalability of DCMIX
allows a model training workflow were channel importance is -- per default -- applied, such that a scientist gets immediate feedback about
where the classification relevant information is coming from, and whether it correlates with the known understanding of the underlying biology.\\
DCMIX scales very well with the number of channels due to the addition of only one additional parameter per additional channel.
This is particularly interesting for hyper spectral applications where hundreds of channels exist (e.g. in remote sensing) --
a highly interesting application area for subsequent studies.\\
In addition, DCMIX allows for any arbitrary downstream network which can be fine-tuned / designed for other applications than fluorescence imaging.\\
DCMIX applies currently a simple addition channel mixing strategy to estimate channel importance without losing any classification performance
(see model performance in Table \ref{tab:pred:mnist} and \ref{tab:pred:rxrx}). In principle several other channel blending methods exist, e.g. difference, multiplication
or luminosity.
Due to the flexibility of DCMIX these other mixing strategies can be easily integrated.
Several studies already show the applicability of complex multi-spectral channel blending for visualization and classification in remote sensing \cite{kokalj_why_2019,jordanova_improved_2023}.

\section{Conclusion}
In this work, we present a novel lightweight framework, DCMIX, which estimates channel importance of fluoresce images based on image blending. This empowers us to estimate phenotype-focused interpretations in a simple yet effective manner. Our experimental results demonstrate that the channel importance scores uncovered by DCMIX are both biologically supported and in line with competitive state-of-the-art approaches on MNIST and RXRX1 datasets. Concurrently, DCMIX is more effective in terms of runtime and scaleable to an arbitrary number of channels without scarifying the model performance.  \\

\noindent \textbf{Limitations.} We discuss the limitations of our approach in the following two aspects. (1) The weights of DCMIX which determine the channel importance are solely a proxy and do not explain the absolute importance between channels. (2) DCMIX is based on image blending and hence only supporting image-based datasets. For future work, we plan to investigate how DCMIX can be extended to other data modalities.

\bibliographystyle{splncs04}
\bibliography{098-main.bib}

\end{document}